\documentclass{article}\usepackage{spconf,amsmath,graphicx,cite}     

\usepackage{graphics} 

\def\onen #1{\left\|#1\right\|_1}

\def\frobn #1{\left\|#1\right\|_{\text{F}}}

\def\sgn #1{\text{sgn}#1}

\def\abs #1{\left|#1\right|}
\def\inp #1{\left\langle#1\right\rangle}

\def\m #1{\boldsymbol{#1}}

\def\cD{\mathcal{D}}
\def\cF{\mathcal{F}}

\def\cI{\mathcal{I}}

\def\cP{\mathcal{P}}

\def\cR{\mathcal{R}}
\def\cT{\mathcal{T}}

\def\cW{\mathcal{W}}

\def\bee{\begin{equation}}
\def\ene{\end{equation}}

\def\beq{\begin{eqnarray}}
\def\enq{\end{eqnarray}}

\def\equ #1{\begin{equation}#1\end{equation}}

\def\sbra #1{\left(#1\right)}
\def\mbra #1{\left[#1\right]}
\def\lbra #1{\left\{#1\right\}}

\title{Sparse MRI for Motion Correction}
\name{Zai Yang$^*$, Cishen Zhang$^\dag$, and Lihua Xie$^*$, Fellow, IEEE
\thanks{The research was in part supported by the National Research Foundation of China under grant NSFC 61120106011.}}
\address{$^*$EXQUISITUS, Centre for E-City, School of Electrical and Electronic Engineering,\\
Nanyang Technological University, 639798, Singapore\\
$^\dag$Faculty of Engineering and Industrial Sciences, Swinburne University of Technology, \\Hawthorn VIC 3122, Australia}

\begin{document}
\maketitle

\begin{abstract}
MR image sparsity/compressibility has been widely exploited for imaging acceleration with the development of compressed sensing. A sparsity-based approach to rigid-body motion correction is presented for the first time in this paper. A motion is sought after such that the compensated MR image is maximally sparse/compressible among the infinite candidates. Iterative algorithms are proposed that jointly estimate the motion and the image content. The proposed method has a lot of merits, such as no need of additional data and loose requirement for the sampling sequence. Promising results are presented to demonstrate its performance.
\end{abstract}



\section{Introduction}
Imaging artifacts due to patient motion remain a major challenge in many MRI applications. In this paper, we consider MR image reconstruction from $k$-space data corrupted by 2D translational motion. The problem is ill-posed since there exist infinitely many image candidates and possible motions which result in the same $k$-space data. The rigid-body motion correction problem has been an active research topic for a long time. Existing methods resolve the problem by acquiring more $k$-space data and/or adopting a specific sampling sequence. Early works \cite{medley1991improved,zoroofi1995improved} assume that the region of interest (ROI) is {\em a priori} known and do motion correction by borrowing the idea of phase retrieval \cite{marchesini2007unified}. A relatively small ROI is helpful for the motion correction and is obtained by oversampling in the frequency domain, or equivalently, imposing an enlarged field of view (FOV) in the image domain. Navigator-based methods \cite{kadah2004floating, lin2010motion} have been popular in the past decade which resolve the ill-posed problem by acquiring extra $k$-space lines named as ``navigator''. The motion detected between the navigator and a motion-free reference is used for data correction. Translational motion is studied in \cite{kadah2004floating} while \cite{lin2010motion} also considers rotational motion. In \cite{mendes2009rigid}, the motion is estimated by exploiting correlations between adjacent readout lines by assuming that little motion exists during a single echotrain (the same assumption required in navigator-based methods). To guarantee the existence of the correlations, oversampling is required along the phase-encode direction \cite{mendes2009rigid}.

Compressed sensing (CS) \cite{candes2006near} has recently become a powerful tool for signal processing that aims at reconstructing a high dimensional signal from its low dimensional linear measurements. The inherent reason of its success is that most natural signals can be sparsely represented in some basis, rendering that a small number of linear measurements are sufficient for the signal recovery. Since medical images are sparse/compressible, for example, in a wavelet domain, applications of CS to MRI have been vastly reported in the literature (e.g., \cite{lustig2007sparse}) for accelerating the scanning process by downsampling in the $k$-space.

In this paper, we study the motion correction problem by exploiting the MR image sparsity. To do this, we formulate the motions as unknown parameters in the sampling system. Rather than acquiring partial $k$-space data in the motion-free environment for imaging acceleration, we attempt to estimate the system parameters jointly with the image reconstruction process by exploiting the redundancy in the full-sampled $k$-space data. The idea is inspired by our recent work \cite{yang2012robustly} where it is shown that uncertain system parameters can be accurately estimated along with the sparse signal in one particular situation. The proposed method relies on the assumption that little motion occurs during a single readout line, which can be approximately satisfied in practice and is greatly relaxed in comparison with that in \cite{kadah2004floating,mendes2009rigid}. Unlike existing navigator-based methods, it does not require additional $k$-space data. The method is justified using a Shepp-Logan phantom and simulated human brain data.

\section{Problem formulation of motion correction}
Consider that a translational motion $\m{\beta}_{\m{k}}$ occurs when acquiring the $k$-space measurement at $\m{k}$. So it holds at the moment
$\overline{m}\sbra{\m{r}}=m^o\sbra{\m{r}-\m{\beta}_{\m{k}}}$,
where $\m{m}^o$ is the MR image of interest, $\overline{\m{m}}$ is the translated image and $\m{r}$ denotes the coordinate in the image domain. According to the relationship between an MR image and its $k$-space measurements, it holds
\equ{\begin{split}\overline{M}\sbra{\m{k}}
&=\iint \overline{m}\sbra{\m{r}}e^{-i2\pi\inp{\m{k},\;\m{r}}}d\m{r}\\
&=e^{-i2\pi\inp{\m{k},\;\m{\beta}_{\m{k}}}}M^o\sbra{\m{k}},
\end{split} \label{formu:kspace_motion}}
where $\m{M}^o=\cF\m{m}^o$ and $\overline{\m{M}}$ denote the motion-free and motion-corrupted $k$-space data, respectively, with $\cF$ denoting the 2D Fourier transform operator. It is shown in (\ref{formu:kspace_motion}) that the translational motion leads to a phase error and unaltered amplitude. To represent (\ref{formu:kspace_motion}) more compactly, let $\m{\beta}=\mbra{\cdots,\m{\beta}_{\m{k}_j},\cdots}^T$ and then write (\ref{formu:kspace_motion}) into
\equ{\overline{\m{M}}=\cT_{\m{\beta}}\cF\m{m}^o, \label{formu:image_data_equation}}
where $\cT_{\m{\beta}}$ denotes a linear operator caused by the translational motion $\m{\beta}$ and is referred to as the translational operator hereafter. In particular, let $\m{\Lambda}_{\m{\beta}}$ be a matrix of the same dimension as $\overline{\m{M}}$ with its element $\Lambda_{\m{\beta}}\sbra{\m{k}}=e^{-i2\pi\inp{\m{k},\;\m{\beta}_{\m{k}}}}$. Then $\cT_{\m{\beta}}\m{M}^o=\m{\Lambda}_{\m{\beta}}\odot\m{M}^o$ where $\odot$ denotes the Hadamard product. In addition, it holds that $\cT_{\m{\beta}}^{-1}=\cT_{-\m{\beta}}$ and $\cT_{\m{0}}=\cI$ where $\cI$ denotes the identity operator. For a particular motion $\m{\beta}$, the linear system of equations in (\ref{formu:image_data_equation}) relates the MR image $\m{m}^o$ of interest and the acquired $k$-space data $\overline{\m{M}}$. Our objective is to reconstruct the image $\m{m}^o$ and possibly the motion $\m{\beta}$ given the corrupted $k$-space data $\overline{\m{M}}$.

The challenge is that the problem of recovering $\m{m}^o$ from $\overline{\m{M}}$ is ill-posed since there exist infinite number of solutions to (\ref{formu:image_data_equation}). In particular, for any $\m{\beta}$ there exists a solution $\m{m}=\cF^{-1}\cT_{-\m{\beta}}\overline{\m{M}}$ such that (\ref{formu:image_data_equation}) holds. To choose the correct one among the infinite candidates, as a result, additional information has to be exploited.

\section{Sparsity-driven Motion Correction}
\subsection{Sparsity-based formulation} \label{sec:CS_formulation}
We first consider that the MR image is directly constructed from the corrupted $k$-space data $\overline{\m{M}}$ without considering the translational motion. Then, the constructed image is
\equ{\widehat{\m{m}}=\cF^{-1}\overline{\m{M}} =\cF^{-1}\sbra{\m{\Lambda}_{\m{\beta}}\odot\m{M}^o} =\sbra{\cF^{-1}\m{\Lambda}_{\m{\beta}}}\otimes\m{m}^o, \label{formu:mhat}}
where $\otimes$ denotes the circular convolution operation. So the obtained image is the true image $\m{m}^o$ after a convolution with $\cF^{-1}\m{\Lambda}_{\m{\beta}}$. That explains how the imaging artifacts come from the translational motion. Due to the imaging artifacts, it is natural to conjecture that the translational motion will reduce the sparsity/compressibility of MR images, i.e., the true image is the maximally sparse/compressible solution (under an appropriate basis) to (\ref{formu:image_data_equation}). Examples of a Shepp-Logan phantom and simulated human brain are presented in Fig. \ref{Fig:results}. In comparison with the motion-free images (in col 1), severe artifacts are present in the motion-corrupted ones (in col 2). Numerically, it can be shown that the $\ell_1$ norms (a commonly used sparsity metric) of the motion-corrupted images (under an Haar wavelet basis) are larger than those of the motion-free ones.

Based on the conjecture above, we propose to reconstruct the motion-free MR image using the maximally sparse solution that satisfies the data consistency constraint (\ref{formu:image_data_equation}), i.e., by solving a basis pursuit (BP) like optimization problem:
\equ{\begin{split}
&\min_{\m{m},\m{\beta}} \onen{\cW\m{m}}\\
&\text{ subject to } \cT_{\m{\beta}}\cF\m{m}=\overline{\m{M}} \text{ and } \m{\beta}\in\cD_{\m{\beta}},\end{split} \label{formu:BP}}
where $\cW$ is a sparsifying operator (e.g., a wavelet transform), $\onen{\cdot}$ denotes the $\ell_1$ norm (sum of amplitude of all elements) that is commonly used to promote sparsity and $\cD_{\m{\beta}}$ denotes the domain of $\m{\beta}$.
Given an appropriate constant $C$, an equivalent (LASSO-like) formulation of (\ref{formu:BP}) is
\equ{\begin{split}
&\min_{\m{m},\m{\beta}} \frobn{\overline{\m{M}}-\cT_{\m{\beta}}\cF\m{m}},\\
&\text{ subject to } \onen{\cW\m{m}}\leq C  \text{ and } \m{\beta}\in\cD_{\m{\beta}},\end{split} \label{formu:Lasso}}
where $\frobn{\cdot}$ denotes the Frobenius norm and $C<\onen{\cW\cF^{-1}\overline{\m{M}}}$ is a constant that needs to be tuned in practice and can be set to $C=\onen{\cW\m{m}^o}$ in an ideal case. The motions $\m{\beta}_{\m{k}}$ with respect to the coordinate $\m{k}$ are not independent. For example, the motion should be piecewise smooth when sorted chronologically. In this paper, we assume that the same motion is shared among every readout line, which can be approximately satisfied in practice and greatly reduces the number of unknown parameters. Note that the assumption is greatly relaxed comparing to that in \cite{kadah2004floating,mendes2009rigid} where it is assumed that little motion exists during every echotrain (each comprising several readout lines). Moreover, $\m{\beta}$ should be properly bounded in practice. These {\em a priori} knowledge narrows the selection of $\m{\beta}$ and defines $\cD_{\m{\beta}}$. To the best of our knowledge, (\ref{formu:BP}) and (\ref{formu:Lasso}) present the first sparsity-based formulations for the MRI motion correction problem though CS has been widely applied to MRI for imaging acceleration.

\subsection{An alternating algorithm} \label{sec:iterative_alg}
Due to the nonconvexity with respect to $\m{\beta}$, problem (\ref{formu:Lasso}) is nonconvex. We propose an alternating algorithm that iteratively carries out the following three steps starting with $j=0$ and $\m{\beta}^{\sbra{0}}=\m{0}$:
\begin{itemize}
 \item[1)] solving
      \equ{\begin{split}\widetilde{\m{m}}^{\sbra{j+1}} =
      &\arg\min_{\m{m}} \frobn{\overline{\m{M}}-\cT_{\m{\beta}^{\sbra{j}}}\cF\m{m}},\\
      &\text{ subject to } \onen{\cW\m{m}}\leq C. \end{split}\label{formu:solvem}}
 \item[2)] solving
      \equ{\begin{split}\m{\beta}^{\sbra{j+1}}=
      &\arg\min_{\m{\beta}} \frobn{\overline{\m{M}}-\cT_{\m{\beta}}\cF\widetilde{\m{m}}^{\sbra{j+1}}},\\
      &\text{ subject to } \m{\beta}\in\cD_{\m{\beta}}. \end{split} \label{formu:solvebeta}}
 \item[3)] letting $j\leftarrow j+1$.
\end{itemize}

To interpret the algorithm above, denote two sets $S_1=\lbra{\m{m}: \onen{\cW\m{m}}\leq C}$ and $S_2=\lbra{\m{m}: \overline{\m{M}}=\cT_{\m{\beta}}\cF\m{m}, \m{\beta}\in\cD_{\m{\beta}}}$. We refer to $S_1$ as the {\em sparse domain} and $S_2$ as the {\em Fourier domain}, respectively. Denote $\cP_1$ and $\cP_2$ projections onto $S_1$ and $S_2$, respectively. Let $\m{m}^{\sbra{j}}= \cF^{-1}\cT_{-\m{\beta}^{\sbra{j}}}\overline{\m{M}}$. Then it holds $\widetilde{\m{m}}^{\sbra{j+1}}=\cP_1\m{m}^{\sbra{j}}$ following from {\em Step 1}. At {\em Step 2}, the equality $\m{m}^{\sbra{j+1}}=\cP_2\widetilde{\m{m}}^{\sbra{j+1}}$ holds. So the algorithm above is equivalent to the recursion
\equ{\m{m}^{\sbra{j+1}}=\cP_2\cP_1\m{m}^{\sbra{j}}, \label{formu:ER}}
starting with $\m{m}^{\sbra{0}}=\widehat{\m{m}}$ where $\widehat{\m{m}}$ is as defined in (\ref{formu:mhat}).
Since $\m{m}^{\sbra{0}}\in S_2$, the recursion attempts to find a point in $S_2$ nearest to $S_1$, or equivalently, the maximally sparse image that is consistent with the $k$-space observation. The constant $C$ is generally unavailable in advance and needs to be tuned in the algorithm. A simple method is to set $C$ such that the obtained image has the least $\ell_1$ norm in the sparsifying domain (which is consistent with our objective). 
$\cP_1$ (i.e., to solve (\ref{formu:solvem})) is easy to compute. The next subsection is devoted to computing $\cP_2$.

\subsection{Approximate computation of $\cP_2$}
The computation of $\cP_2$ is equivalent to solving the optimization problem in (\ref{formu:solvebeta}). Under the assumption that a common displacement is shared among each readout line and the motions among different readout lines are independent, the number of unknown variables in $\m{\beta}$ is reduced to twice the number of readout lines (each displacement has 2 degrees of freedom in the 2D case) and the displacement in each line can be separately calculated. However, the exact calculation of $\cP_2$, or equivalently, to obtain the global optimum of (\ref{formu:solvebeta}), is generally infeasible since the Fourier domain $S_2$ is severely nonconvex. The $\m{\beta}$ solution is typically trapped at a local minimum if optimization methods, e.g., a gradient method, are used to solve (\ref{formu:solvebeta}). Inspired by a navigator-based method in \cite{lin2010motion}, we propose a practically efficient algorithm as follows.

In (\ref{formu:solvebeta}), the term $\cF\widetilde{\m{m}}^{\sbra{j}}$, denoted by $\m{M}^{\sbra{j}}$, can be considered as the current estimate of the motion-free $k$-space data $\m{M}^o$ (without accounting for the difference between $\widetilde{\m{m}}^{\sbra{j}}$ and $\m{m}^{\sbra{j}}$). To solve (\ref{formu:solvebeta}) is in fact to estimate the translational motion between the motion-corrupted observation $\overline{\m{M}}$ and the current motion-free estimate $\m{M}^{\sbra{j}}$. So we may consider each readout line of $\overline{\m{M}}$ as a navigator and the associated line of $\m{M}^{\sbra{j}}$ as the reference. Then the motion within each readout line can be estimated using the navigator-based method in \cite{lin2010motion}. We omit the details.
At last, we point out that the algorithm performance can be improved in practice by a modification of the current motion-free $k$-space estimate: $\m{M}^{\sbra{j}}=\abs{\overline{\m{M}}}\odot \sgn\sbra{\cF\widetilde{\m{m}}^{\sbra{j}}}$, instead of using $\m{M}^{\sbra{j}}=\cF\widetilde{\m{m}}^{\sbra{j}}$, where the absolute operator $\abs{\cdot}$ does an element-wise operation. The underlying reason is obvious according to (\ref{formu:kspace_motion}). The translational motion changes only the phase information of the $k$-space data and keeps the amplitude. So intuitively, the modified estimate is closer to the true motion-free data and leads to better performance.

\subsection{Sparse RAAR for motion correction}

According to (\ref{formu:ER}), the algorithm in Subsection \ref{sec:CS_formulation} is implemented by iteratively projecting onto the two sets $S_1$ and $S_2$. It is related to iterative projection algorithms \cite{marchesini2007unified} for phase retrieval, which aims to find an intersection point of two sets and are built upon combining projections onto the two sets in some fashion. In fact, the recursion (\ref{formu:ER}) is exactly the error reduction (ER) algorithm \cite{marchesini2007unified} without accounting for the differences of the sets. One drawback of ER is its slow convergence. In our setting, we want to solve a feasibility problem:
\equ{\text{find } \m{m}\in S_1\cap S_2,}
i.e., to find a point lying in both the sparse domain $S_1$ and the Fourier domain $S_2$. The relaxed averaged alternating reflections (RAAR) algorithm introduced in \cite{luke2005relaxed} has fast convergence speed and stable performance. RAAR starts with some initial point $\m{m}^{\sbra{0}}$ and is defined by the recursion
\equ{\m{m}^{\sbra{j+1}}= \sbra{\frac{\theta}{2}\sbra{\cR_1\cR_2+\cI}+\sbra{1-\theta}\cP_2}\m{m}^{\sbra{j}},}
where $\theta\in\mbra{0,\;1}$ is a constant, and reflectors $\cR_1=2\cP_1-\cI$ and $\cR_2=2\cP_2-\cI$. By defining and computing $\cP_1$ and $\cP_2$ as before, we propose an algorithm for the motion correction problem, named as sparse RAAR (SRAAR), where $\widehat{\m{m}}$ in (\ref{formu:mhat}) is used as the initial point.


\section{Results}
The proposed method is validated using a Shepp-Logan phantom and simulated human brain data obtained from BrainWeb \footnote{http://www.bic.mni.mcgill.ca/brainweb} with both of size $256\times 256$. Continuously varying translational motions among the readout lines are randomly generated. Motion artifacts are introduced by applying varying linear phase shifts to the motion-free $k$-space data according to (\ref{formu:kspace_motion}). An Haar wavelet is selected as the sparsifying transform. SRAAR is applied to the motion-corrupted $k$-space data to reconstruct the MR images with the setting $\theta=0.9$. SRAAR is terminated after a fixed number of iterations.

\begin{figure}
\centering
  \includegraphics[width=3.5in]{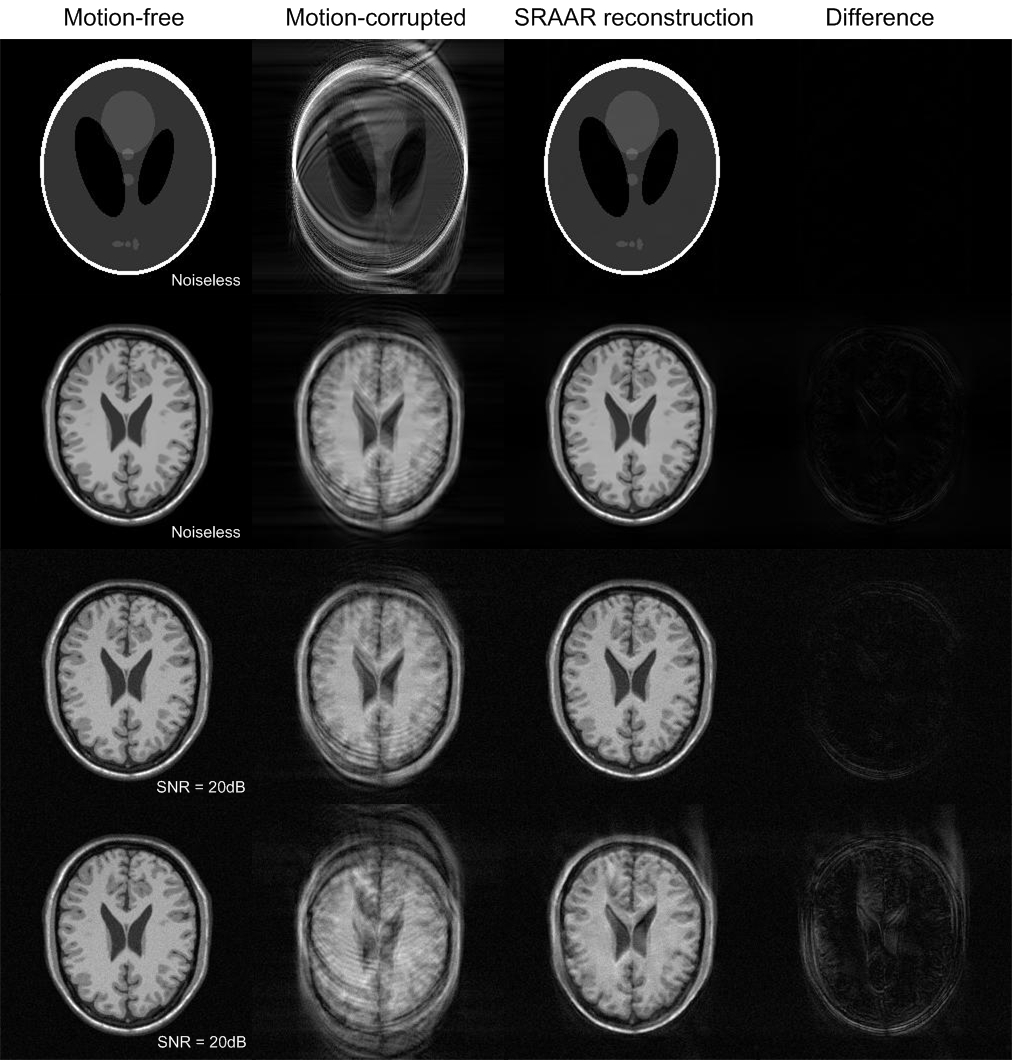}
\centering
\caption{Simulation results of sparsity-based motion correction on a Shepp-Logan phantom (row 1, 100 iterations), simulated human brain (row 2, 200 iterations), and noisy brain with small motions (row 3, 200 iterations) and large motions (row 4, 400 iterations).} \label{Fig:results}
\end{figure}

Simulation results are presented in Fig. \ref{Fig:results}. For the phantom (row 1), almost exact reconstruction is obtained using SRAAR. A small amount of motions (within 5 pixels along both the readout and phase-encode directions) are studied in rows 2 (without noise) and 3 (noise added) with the human brain. It can be seen that even a small amount of motions may cause severe imaging artifacts. After the sparsity-based motion correction with SRAAR, only few artifacts remain. In the presence of both noises and a large amount of motions (as 3 times large as those in rows 2 and 3), it is shown in row 4 that SRAAR may have difficulties to produce a good result though most artifacts are removed. We note that SRAAR is computationally efficient in general. In the above, each iteration takes about 1s in Matlab v.7.7.0 on a PC with a 3GHz CPU, i.e., each reconstructed image is obtained within few minutes. Moreover, SRAAR can be greatly accelerated by estimating motions in the readout lines in parallel when computing $\cP_2$.

\section{Conclusion}
A first sparsity-based approach to motion correction was introduced in this paper. An efficient algorithm was proposed partially inspired by phase retrieval and existing navigator-based methods. The promising results presented in the paper indicate good potential for practical application of the proposed method. The current work considers only translational motion and encourages further studies of more complicated motions. Reference \cite{yang2012robustly} shows that unknown system parameters can be estimated even in the undersampling case, suggesting that imaging acceleration is also possible in the presence of motions. A shortcoming of the algorithms in this paper is the need of the tuning of $C$. A recent improvement in the algorithm design can be found in \cite{yang2013robust}.

\bibliographystyle{IEEEbib}


\end{document}